\documentclass[10pt,twocolumn,letterpaper]{article}

\usepackage{cvpr}
\usepackage{times}
\usepackage{epsfig}
\usepackage{graphicx}
\usepackage{amsmath}
\usepackage{amssymb}

\usepackage{diagbox}
\usepackage{color}
\usepackage{xcolor}
\usepackage{multirow}
\usepackage{makecell}
\usepackage{enumitem}

\usepackage{dcolumn}
\newcolumntype{d}[1]{D{.}{.}{#1}}

\usepackage[british,english,american]{babel}

\newcolumntype{L}[1]{>{\raggedright\let\newline\\\arraybackslash\hspace{0pt}}m{#1}}
\newcolumntype{C}[1]{>{\centering\let\newline\\\arraybackslash\hspace{0pt}}m{#1}}
\newcolumntype{R}[1]{>{\raggedleft\let\newline\\\arraybackslash\hspace{0pt}}m{#1}}

\definecolor{citecolor}{RGB}{34,139,34}
\usepackage[pagebackref=true,breaklinks=true,letterpaper=true,citecolor=citecolor,colorlinks,bookmarks=false]{hyperref}

\newcommand{\app}{\raise.17ex\hbox{$\scriptstyle\sim$}}

\definecolor{demphcolor}{RGB}{144,144,144}
\newcommand{\demph}[1]{\textcolor{demphcolor}{#1}}

\newlength\savewidth\newcommand\shline{\noalign{\global\savewidth\arrayrulewidth
  \global\arrayrulewidth 1pt}\hline\noalign{\global\arrayrulewidth\savewidth}}

\makeatletter\renewcommand\paragraph{\@startsection{paragraph}{4}{\z@}
  {.5em \@plus1ex \@minus.2ex}{-.5em}{\normalfont\normalsize\bfseries}}\makeatother

 \cvprfinalcopy 

\def\cvprPaperID{****} 

\ifcvprfinal\fi
\begin{document}

\title{Feature Denoising for Improving Adversarial Robustness}

\author{
Cihang Xie\textsuperscript{1,2}\footnotemark \qquad
Yuxin Wu\textsuperscript{2} \qquad
Laurens van der Maaten\textsuperscript{2} \qquad
Alan Yuille\textsuperscript{1} \qquad
Kaiming He\textsuperscript{2} \vspace{.3em}\\
\textsuperscript{1}Johns Hopkins University \qquad\qquad \textsuperscript{2}Facebook AI Research
\vspace{-.5em}
}

\maketitle
 \renewcommand*{\thefootnote}{\fnsymbol{footnote}}
 \setcounter{footnote}{1}
 \footnotetext{Work done during an internship at Facebook AI Research.}
 \renewcommand*{\thefootnote}{\arabic{footnote}}
 \setcounter{footnote}{0}



\begin{abstract}
\vspace{-0.3cm}
Adversarial attacks to image classification systems present challenges to convolutional networks and opportunities for understanding them.
This study suggests that adversarial perturbations on images lead to noise in the features constructed by these networks. Motivated by this observation, we develop new network architectures that increase adversarial robustness by performing feature denoising. Specifically, our networks contain blocks that denoise the features using non-local means or other filters; the entire networks are trained end-to-end. When combined with adversarial training, our feature denoising networks substantially improve the state-of-the-art in adversarial robustness in both white-box and black-box attack settings.
On ImageNet, under 10-iteration PGD white-box attacks where prior art has 27.9\% accuracy, 
our method achieves 55.7\%; even under extreme 2000-iteration PGD white-box attacks, our method secures 42.6\% accuracy.
Our method was ranked first in Competition on Adversarial Attacks and Defenses (CAAD) 2018 --- it achieved 50.6\% classification accuracy on a secret, ImageNet-like test dataset against 48 unknown attackers, surpassing the runner-up approach by $\app$10\%. Code is available at \url{https://github.com/facebookresearch/ImageNet-Adversarial-Training}.
\end{abstract}


\begin{figure}[t]
\centering
\includegraphics[width=0.75\linewidth]{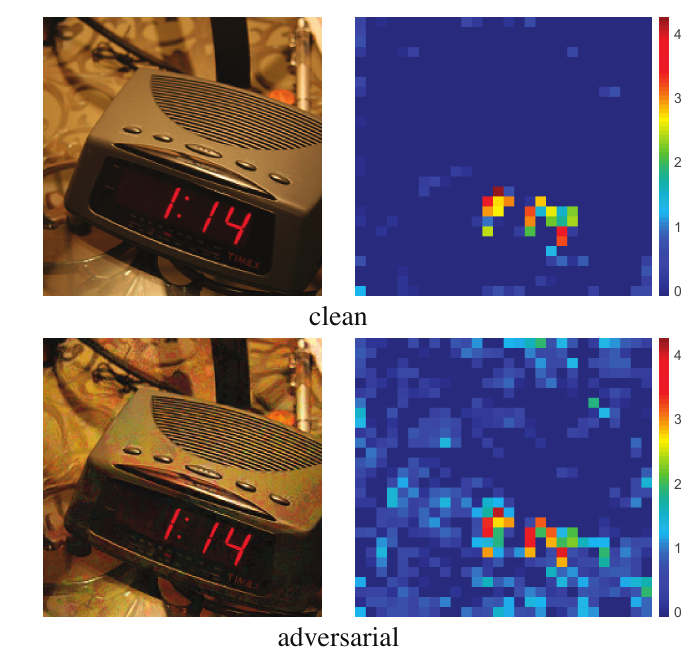}
\vspace{-.2em}
\caption{Feature map in the res$_3$ block of an ImageNet-trained ResNet-50 \cite{He2016} applied on a clean image (top) and on its adversarially perturbed counterpart (bottom). The adversarial  perturbation was produced using PGD~\cite{Madry2018} with maximum perturbation $\epsilon\!=\!$ 16 (out of 256). In this example, the adversarial image is incorrectly recognized as ``space heater''; the true label is ``digital clock''.}
\label{fig:teaser}
\vspace{-1.2em}
\end{figure}

\vspace{-0.5em}
\section{Introduction}

Adversarial attacks to image classification systems \cite{Szegedy2014} add small perturbations to images that lead these systems into making incorrect predictions. While the perturbations are often imperceptible or perceived as small ``noise'' in the image, these attacks are highly effective against even the most successful convolutional network based systems~\cite{Krizhevsky2012,LeCun1989}. The success of adversarial attacks leads to security threats in real-world applications of convolutional networks, but equally importantly, it demonstrates that these networks perform computations that are dramatically different from those in human brains.

\begin{figure*}[t]
\centering
\vspace{-.5em}
\includegraphics[width=1.\linewidth]{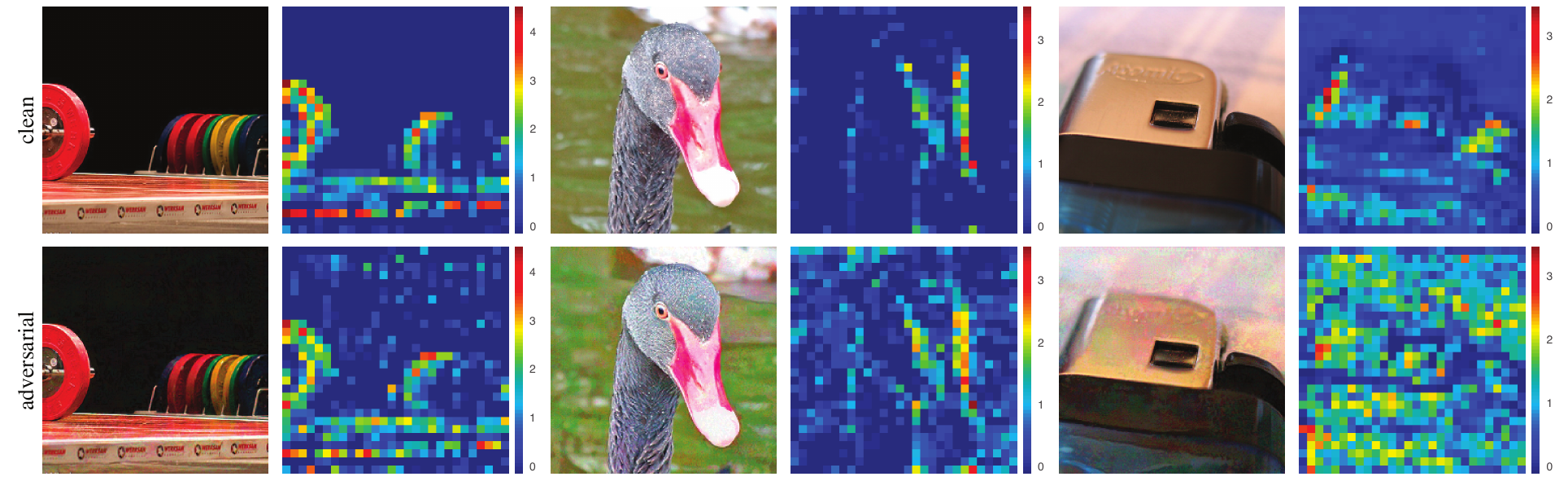}
\vspace{-1em}
\caption{More examples similar to Figure~\ref{fig:teaser}. We show feature maps corresponding to clean images (top) and to their adversarial perturbed versions (bottom). The feature maps for each pair of examples are from the same channel of a res$_3$ block in the same ResNet-50 trained on clean images. The attacker has a maximum perturbation $\epsilon=16$ in the pixel domain.}
\label{fig:noise}
\vspace{-1.5em}
\end{figure*}

Figure~\ref{fig:teaser} shows a randomly selected feature map of a ResNet \cite{He2016} applied on a clean image (top) and on its adversarially perturbed counterpart (bottom). The figure suggests that adversarial perturbations, while small in the pixel space, lead to very substantial ``noise'' in the feature maps of the network. Whereas the features for the clean image appear to focus primarily on semantically informative content in the image, the feature maps for the adversarial image are activated across semantically irrelevant regions as well. Figure~\ref{fig:noise} displays more examples with the same pattern.

Motivated by this observation, we explore \emph{feature denoising} approaches to improve the robustness of convolutional networks against adversarial attacks.
We develop new convolutional network architectures equipped with building blocks designed to denoise feature maps. Our networks are trained end-to-end on adversarially generated samples, allowing them to learn to reduce feature-map perturbations. 

Empirically, we find that the best performance is achieved by networks using \emph{non-local means} \cite{Buades2005} for feature denoising, leading to models that are related to self-attention \cite{Vaswani2017} and non-local networks \cite{Wang2018}. Our ablation studies show that using \emph{mean filters}, \emph{median filters}, and \emph{bilateral filters} \cite{Tomasi1998} for feature denoising also improves adversarial robustness, suggesting that feature denoising is a good design principle.

Our models outperform the state-of-the-art in adversarial robustness against highly challenging \emph{white-box} attacks on ImageNet \cite{Russakovsky2015}.
Under 10-iteration PGD attacks \cite{Madry2018}, we report 55.7\% classification accuracy on ImageNet, largely surpassing the prior art's 27.9\% \cite{Kannan2018} with the same attack protocol.
Even when faced with extremely challenging \emph{2000-iteration} PGD attacks that have not been explored in other literature, our model achieves 42.6\% accuracy. Our ablation experiments also demonstrate that feature denoising consistently improves adversarial defense results in white-box settings.

Our networks are also highly effective under the \emph{black-box} attack setting.
The network based on our method won the defense track in the recent Competition on Adversarial Attacks and Defenses (CAAD) 2018, achieving 50.6\% accuracy against 48 unknown attackers, under a strict ``\emph{all-or-nothing}" criterion. This is an 10\% absolute (20\% relative) accuracy increase compared to the CAAD 2018 runner-up model.
We also conduct ablation experiments in which we defend against the five strongest attackers from CAAD 2017 \cite{Kurakin2018}, demonstrating the potential of feature denoising.

\section{Related Work}

\emph{Adversarial training} \cite{Goodfellow2015,Kannan2018,Madry2018} defends against adversarial perturbations by training networks on adversarial images that are generated on-the-fly during training. Adversarial training constitutes the current state-of-the-art in adversarial robustness against white-box attacks; we use it to train our networks. \emph{Adversarial logit paring} (ALP) \cite{Kannan2018} is a type of adversarial training that encourages the logit predictions of a network for a clean image and its adversarial counterpart to be similar. ALP can be interpreted as ``denoising" the logit predictions for the adversarial image, using the logits for the clean image as the ``noise-free" reference.

Other approaches to increase adversarial robustness include \emph{pixel denoising}. Liao \etal \cite{Liao2018} propose to use high-level features to guide the pixel denoiser; in contrast, our denoising is applied directly on features.
Guo \etal \cite{Guo2018} transform the images via non-differentiable image preprocessing, like image quilting \cite{Efros2001}, total variance minimization \cite{Rudin1992}, and quantization. While these defenses may be effective in black-box settings, they can be circumvented in white-box settings because the attacker can approximate the gradients of their non-differentiable computations \cite{Athalye2018}. In contrast to \cite{Guo2018}, our feature denoising models are \emph{differentiable}, but are still able to improve adversarial robustness against very strong white-box attacks.

\section{Feature Noise}\label{sec:noise}
Adversarial images are created by adding perturbations to images, constraining the magnitude of perturbations to be small in terms of a certain norm (\eg, $L_\infty$ or $L_2$).
The perturbations are assumed to be either imperceptible by humans, or perceived as noise that does not impede human recognition of the visual content. Although the perturbations are constrained to be small at the \emph{pixel level}, no such constraints are imposed at the \emph{feature level} in convolutional networks. Indeed, the perturbation of the features induced by an adversarial image gradually increases as the image is propagated through the network \cite{Liao2018, Guo2018}, and non-existing activations in the feature maps are hallucinated. In other words, the transformations performed by the layers in the network exacerbate the perturbation, and the hallucinated activations can overwhelm the activations due to the true signal, which leads the network to make wrong predictions.

\begin{figure}[t]
\centering
\vspace{-.5em}
\includegraphics[width=\linewidth]{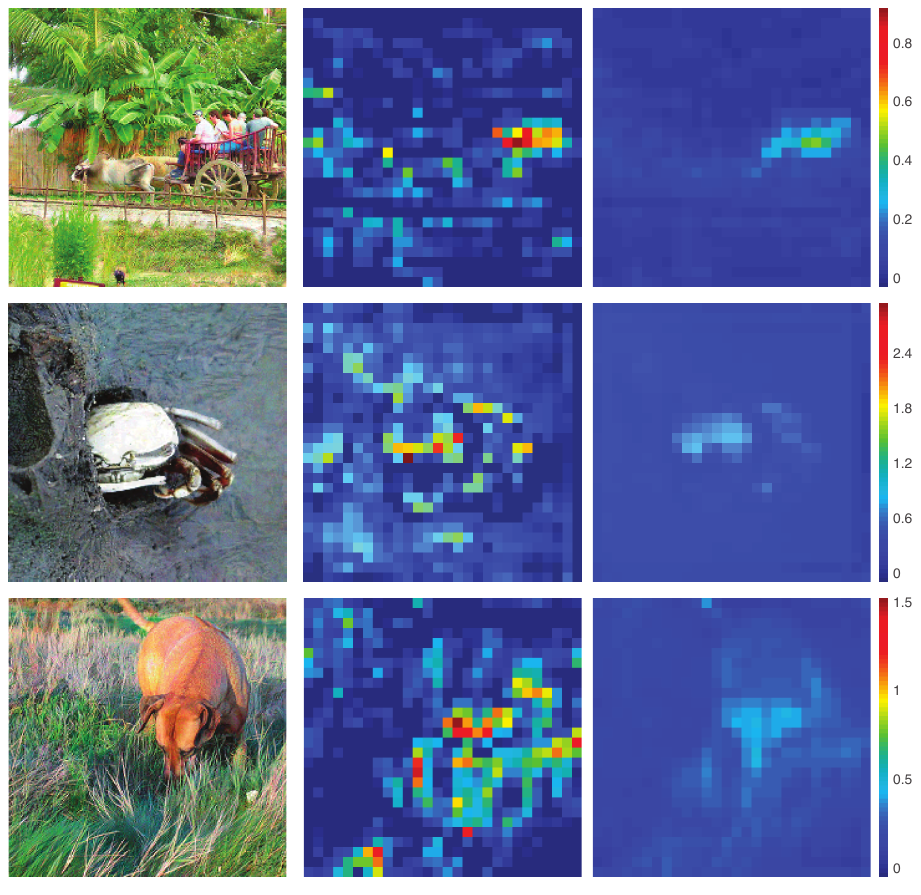}
\vspace{-1.5em}
\caption{Adversarial images and their feature maps \emph{before} (left) and \emph{after} (right) the \emph{denoising operation} (blue box in Figure~\ref{fig:block}).
Here each pair of feature maps are from the same channel of a res$_3$ block in the same adversarially trained ResNet-50 equipped with (Gaussian) non-local means denoising blocks. The attacker has a maximum perturbation $\epsilon\!=\!16$ for each pixel.}
\label{fig:denoise}
\vspace{-1em}
\end{figure}

We qualitatively demonstrate these characteristics of adversarial images by visualizing the feature maps they give rise to. Given a clean image and its adversarially perturbed counterpart, we use the same network (here, a ResNet-50 \cite{He2016}) to compute its activations in the hidden layers. Figure~\ref{fig:teaser} and~\ref{fig:noise} show typical examples of the same feature map on clean and adversarial images, extracted from the middle of the network (in particular, from a res${_3}$ block). These figures reveal that the feature maps corresponding to adversarial images have activations in regions without relevant visual content that resemble feature noise. Assuming that strong activations indicate the presence of semantic information about the image content (as often hypothesized \cite{Zeiler2014}), the activations that are hallucinated by adversarial images reveal why the model predictions are altered.

In this study, we attempt to address this problem by feature denoising. In Figure~\ref{fig:denoise}, we visualize the feature maps of adversarial images, right before and right after a feature denoising operation (see the next section for details). The figure shows that feature denoising operations can successfully suppress much of the noise in the feature maps, and make the responses focus on visually meaningful content. In the next sections, we present empirical evidence showing that models that perform feature denoising operations, indeed, improve adversarial robustness.

Before we move on to describing our methods, we note that although the feature noise can be easily observed \emph{qualitatively}, it is difficult to \emph{quantitatively} measure this noise. We found it is nontrivial to compare feature noise levels between different models, in particular, when the network architecture and/or training methods (standard or adversarial) change. \Eg, adding a denoising block in a network, end-to-end trained, tends to change the magnitudes/distributions of all features. Nevertheless, we believe the observed \emph{noisy} appearance of features reflects a real phenomenon associated with adversarial images.

\section{Denoising Feature Maps}\label{sec:denoise}

Motivated by the empirical observations above, we propose to improve adversarial robustness by adding denoising blocks at intermediate layers of convolutional networks. The denoising blocks are trained jointly with all layers of the network in an end-to-end manner using adversarial training. The end-to-end adversarial training allows the resulting networks to (partly) eliminate feature map noise that is data-dependent, \ie, noise that is generated by the attacker. It also naturally handles the noise across multiple layers by considering how changes in earlier layers may impact the feature/noise distributions of later layers.

Empirically, we find that the best-performed denoising blocks are inspired by self-attention transformers \cite{Vaswani2017} that are commonly used in machine translation and by non-local networks \cite{Wang2018} that are used for video classification. In this study, we focus on the design of denoising blocks and study their denoising effects. Besides non-local means, we also experiment with simpler denoising operations such as bilateral filtering, mean filtering, and median filtering inside our convolutional networks.

\begin{figure}[t]
\centering
\includegraphics[height=3.5cm]{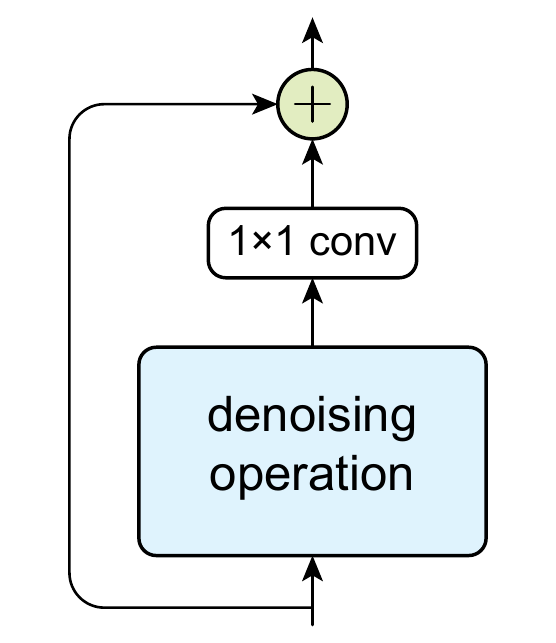}
\caption{A generic denoising block. It wraps the denoising operation (\eg, non-local means, bilateral, mean, median filters) with a 1$\times$1 convolution and an identity skip connection \cite{He2016}.}
\label{fig:block}
\end{figure}

\subsection{Denoising Block}

Figure~\ref{fig:block} shows the generic form of our denoising block. The input to the block can be any feature layer in the convolutional neural network. The denoising block processes the input features by a \emph{denoising operation}, such as non-local means or other variants. The denoised representation is first processed by a 1$\times$1 convolutional layer, and then added to the block's input via a residual connection \cite{He2016}.\footnote{In our terminology, a ``denoising \emph{operation}'' refers to the computation that only performs denoising (blue box in Figure~\ref{fig:block}), and a ``denoising \emph{block}'' refers to the entire block (all of Figure~\ref{fig:block}).}

The design in Figure~\ref{fig:block} is inspired by self-attention \cite{Vaswani2017} and non-local blocks \cite{Wang2018}. However, only the non-local means \cite{Buades2005} operation in the denoising block is actually doing the denoising; the 1$\times$1 convolutions and the residual connection are mainly for feature combination. While various operations can suppress \emph{noise}, they can also impact \emph{signal}. The usage of the residual connection can help the network to retain signals, and the tradeoff between removing noise and retaining signal is adjusted by the 1$\times$1 convolution, which is learned end-to-end with the entire network. We will present ablation studies showing that both the residual connection and the 1$\times$1 convolution contribute to the effectiveness of the denoising block. The generic form of the denoising block allows us to explore various denoising operations, as introduced next.

\subsection{Denoising Operations}

We experiment with four different instantiations of the denoising operation in our denoising blocks.

\begin{figure}[t]
\centering
\includegraphics[height=5.5cm]{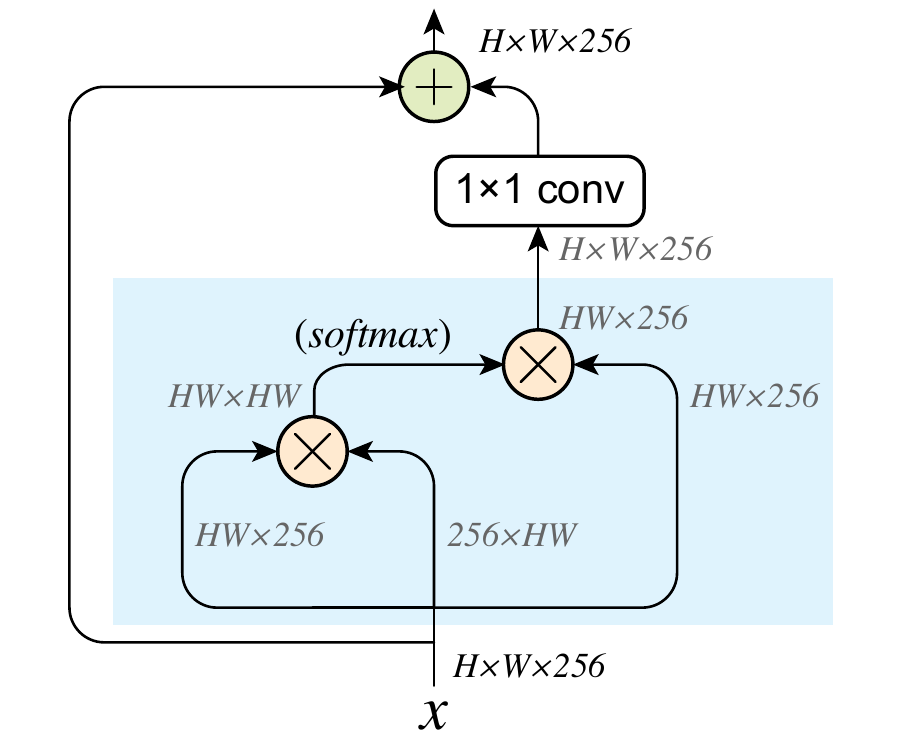}
\caption{A block with \emph{non-local means} as its denoising operation. The blue part illustrates the implementation of non-local means in Eqn.~(\ref{eq:nonlocal}). The shapes of the feature tensors are noted, with corresponding reshaping/transposing performed: here, $H$ and $W$ are the height and width of the feature maps, and we use 256 channels as an example.
If softmax is used, it is the Gaussian version (with appropriate 1$\times$1 convolution embeddings used; omitted in this figure); if softmax is not used, it is the dot product version.}
\vspace{-.1em}
\label{fig:nonlocal_block}
\end{figure}

\paragraph{Non-local means.}
Non-local means \cite{Buades2005} compute a denoised feature map $y$ of an input feature map $x$ by taking a weighted mean of features in all spatial locations $\mathcal{L}$:
\begin{equation}
\label{eq:nonlocal}
y_i = \frac{1}{\mathcal{C}(x)} \sum_{\forall j \in \mathcal{L}} f(x_i, x_j)\cdot x_j,
\end{equation}
where $f(x_i, x_j)$ is a feature-dependent weighting function and $\mathcal{C}(x)$ is a normalization function. 
We note that the weighted average in Eqn.~(\ref{eq:nonlocal}) is over $x_j$, rather than another embedding of $x_j$, unlike \cite{Vaswani2017,Wang2018} --- denoising is directly on the input feature $x$, and the correspondence between the feature channels in $y$ and $x$ is kept.
Following \cite{Wang2018}, we consider two forms: 
\begin{itemize}
\vspace{-.5em}
\item \emph{Gaussian (softmax)} sets $f(x_i, x_j) \!=\! e^{\frac{1}{\sqrt{d}}\theta(x_i)^\text{T} \phi(x_j)}$, where $\theta(x)$ and $\phi(x)$ are two embedded versions of $x$ (obtained by two 1$\times$1 convolutions), $d$ is the number of channels, and $\mathcal{C}\!=\!\sum_{\forall j \in \mathcal{L}} f(x_i,x_j)$. By noticing that $f/\mathcal{C}$ is the \emph{softmax} function, this version is shown in \cite{Wang2018} to be equivalent to the softmax-based, self-attention computation of \cite{Vaswani2017}.

\vspace{-.5em}
\item \emph{Dot product} sets $f(x_i, x_j) = x_i^\text{T} x_j$ and $\mathcal{C}(x) \!=\! N$, where $N$ is the number of pixels in $x$. Unlike the Gaussian non-local means, the weights of the weighted mean do \emph{not} sum up to 1 in dot-product non-local means. However, qualitative evaluations suggest it does suppress noise in the features. Experiments also show this version improves adversarial robustness.
Interestingly, we find that it is \emph{unnecessary} to embed $x$ in the dot-product version of non-local means for the model to work well. This is unlike the Gaussian non-local means, in which embedding is essential. The dot-product version provides a denoising operation with \emph{no} extra parameters (blue box in Figure~\ref{fig:nonlocal_block}).

\end{itemize}

\noindent Figure~\ref{fig:nonlocal_block}, adapted from \cite{Wang2018}, shows the implementation of the denoising block based on non-local means.

\paragraph{Bilateral filter.} It is easy to turn the non-local means in Eqn.~(\ref{eq:nonlocal}) into a ``local mean". Doing so leads to the classical \emph{bilateral filter} \cite{Tomasi1998} that is popular for edge-preserving denoising. Formally, it is defined as:
\begin{equation}
\label{eq:local}
y_i = \frac{1}{\mathcal{C}(x)} \sum_{\forall j \in \Omega(i)} f(x_i, x_j)\cdot x_j.
\end{equation}
This equation only differs from Eqn.~(\ref{eq:nonlocal}) in the neighborhood, $\Omega(i)$, which is a local region (\eg, a 3$\times$3 patch) around pixel $i$. In Eqn.~(\ref{eq:local}), we consider the Gaussian and dot product implementations of the weights as before.

\paragraph{Mean filter.} Perhaps the simplest form of denoising is the mean filter (average pooling with a stride of 1). Mean filters reduce noise but also smooth structures, so it is reasonable to expect them to perform worse than the above weighted means. However, somewhat surprisingly, experiments show that denoising blocks using mean filters as the denoising operation can still improve adversarial robustness.

\paragraph{Median filter.} Lastly, we consider an interesting denoising filter that has rarely been used in deep networks: \emph{median} filtering. The median filter is defined as:
\begin{equation}
\label{eq:median}
y_i = \text{median}\{\forall j \in \Omega(i): x_j \},
\end{equation}
where the median is over a local region, $\Omega(i)$, and is performed separately for each channel.
Median filters are known to be good at removing salt-and-pepper noise and outliers of similar kind. Training convolutional networks that contain median filters is an open problem, but we find experimentally that using median filters as a denoising operation can also improve adversarial robustness.

\vspace{1em}
In summary, our study explores a rich collection of denoising operations. Sec.~\ref{sec:experiments} reports the results for all the denoising operations described above.

\section{Adversarial Training}\label{sec:advtrain}

We show the effectiveness of feature denoising on top of very strong baselines. Our strong experimental results are partly driven by a successful implementation of \emph{adversarial training} \cite{Goodfellow2015,Madry2018}. In this section, we describe our implementation of adversarial training, which is used for training both baseline models and our feature denoising models.

The basic idea of adversarial training \cite{Goodfellow2015,Madry2018} is to train the network on adversarially perturbed images. The adversarially perturbed images can be generated by a given white-box attacker based on the current parameters of the models. We use the Projected Gradient Descent (PGD)\footnote{Publicly available: \fontsize{5.9pt}{1em}\selectfont \url{https://github.com/MadryLab/cifar10_challenge}} \cite{Madry2018} as the white-box attacker for adversarial training. 

\paragraph{PGD attacker.}
PGD is an iterative attacker.
At each iteration, it performs a gradient descent step in the loss function \wrt the image pixel values, based on an adversarially selected output target.
Next, it projects the resulting perturbed images into the feasible solution space --- within a maximum per-pixel perturbation of $\epsilon$ of the clean image (that is, subject to an $L_\infty$ constraint). The hyper-parameters of the PGD attacker during adversarial training are: the maximum perturbation for each pixel $\epsilon \!=\! 16$, the attack step size $\alpha \!=\! 1$, and the number of attack iterations $n \!=\! 30$.
For this PGD in adversarial training, we can initialize the adversarial image by the clean image, or randomly within the allowed $\epsilon$ \cite{Madry2018}. We randomly choose from both initializations in the PGD attacker during adversarial training: 20\% of training batches use clean images to initialize PGD, and 80\% use random points within the allowed $\epsilon$.

\paragraph{Distributed training with adversarial images.} For each mini-batch, we use PGD to generate adversarial images for that mini-batch. Then we perform a one-step SGD on these perturbed images and update the model weights. Our SGD update is based exclusively on adversarial images; the mini-batch contains no clean images.

Because a single SGD update is preceded by $n$-step PGD (with $n \!=\! 30$), the total amount of computation in adversarial training is $\app$$n$$\times$ bigger than standard (clean) training. To make adversarial training practical, we perform distributed training using synchronized SGD on 128 GPUs. Each mini-batch contains 32 images per GPU (\ie, the total mini-batch size is $128 \!\times\! 32 \!=\! 4096$). We follow the training recipe of \cite{Goyal2017}\footnote{Implemented using the publicly available Tensorpack framework~\cite{Wu2016}.} to train models with such large mini-batches. 
On ImageNet, our models are trained for a total of 110 epochs; we decrease the learning rate by 10$\times$ at the 35-th, 70-th, and 95-th epoch. A label smoothing \cite{Szegedy2016a} of 0.1 is used.
The total time needed for adversarial training on 128 Nvidia V100 GPUs is  approximately 38 hours for the baseline ResNet-101 model, and approximately 52 hours for the baseline ResNet-152 model.

\section{Experiments}\label{sec:experiments}

We evaluate feature denoising on the ImageNet classification dataset \cite{Russakovsky2015} that has $\app$1.28 million images in 1000 classes. Following common protocols \cite{Athalye2018,Kannan2018} for adversarial images on ImageNet, we consider \emph{targeted} attacks when evaluating under the white-box settings, where the targeted class is selected uniformly at random; 
targeted attacks are also used in our adversarial training.
We evaluate top-1 \emph{classification accuracy} on the 50k ImageNet validation images that are adversarially perturbed by the attacker (regardless of its targets), also following \cite{Athalye2018,Kannan2018}.

In this paper, adversarial perturbation is considered under $L_\infty$ norm (\ie, maximum difference for each pixel), with an allowed maximum value of $\epsilon$. The value of $\epsilon$ is relative to the pixel intensity scale of 256.

Our baselines are ResNet-101/152 \cite{He2016}.
By default, we add 4 denoising blocks to a ResNet: each is added after the last residual block of res$_2$, res$_3$, res$_4$, and res$_5$, respectively.

\subsection{Against White-box Attacks}
\label{subsec; white-box exp}

Following the protocol of ALP \cite{Kannan2018}, we report defense results against PGD as the white-box attacker.\footnote{We have also evaluated other attackers, including FGSM \cite{Goodfellow2015}, iterative FGSM \cite{Kurakin2017a}, and its momentum variant \cite{Dong2018}.
Similar to \cite{Kannan2018}, we found that PGD is the strongest white-box attacker among them.}
We evaluate with $\epsilon\!=\!$ 16, a challenging case for defenders on ImageNet.

Following~\cite{Madry2018}, the PGD white-box attacker initializes the adversarial perturbation from a random point within the allowed $\epsilon$ cube.
We set its step size $\alpha \!=\! 1$, except for 10-iteration attacks where $\alpha$ is set to $\epsilon / 10 \!=\! 1.6$.
We consider a numbers of PGD attack iterations ranging from 10 to 2000. 

\begin{figure*}[t]
\centering
\vspace{-.5em}
\resizebox{1.\textwidth}{!}{
\includegraphics[height=6.1cm]{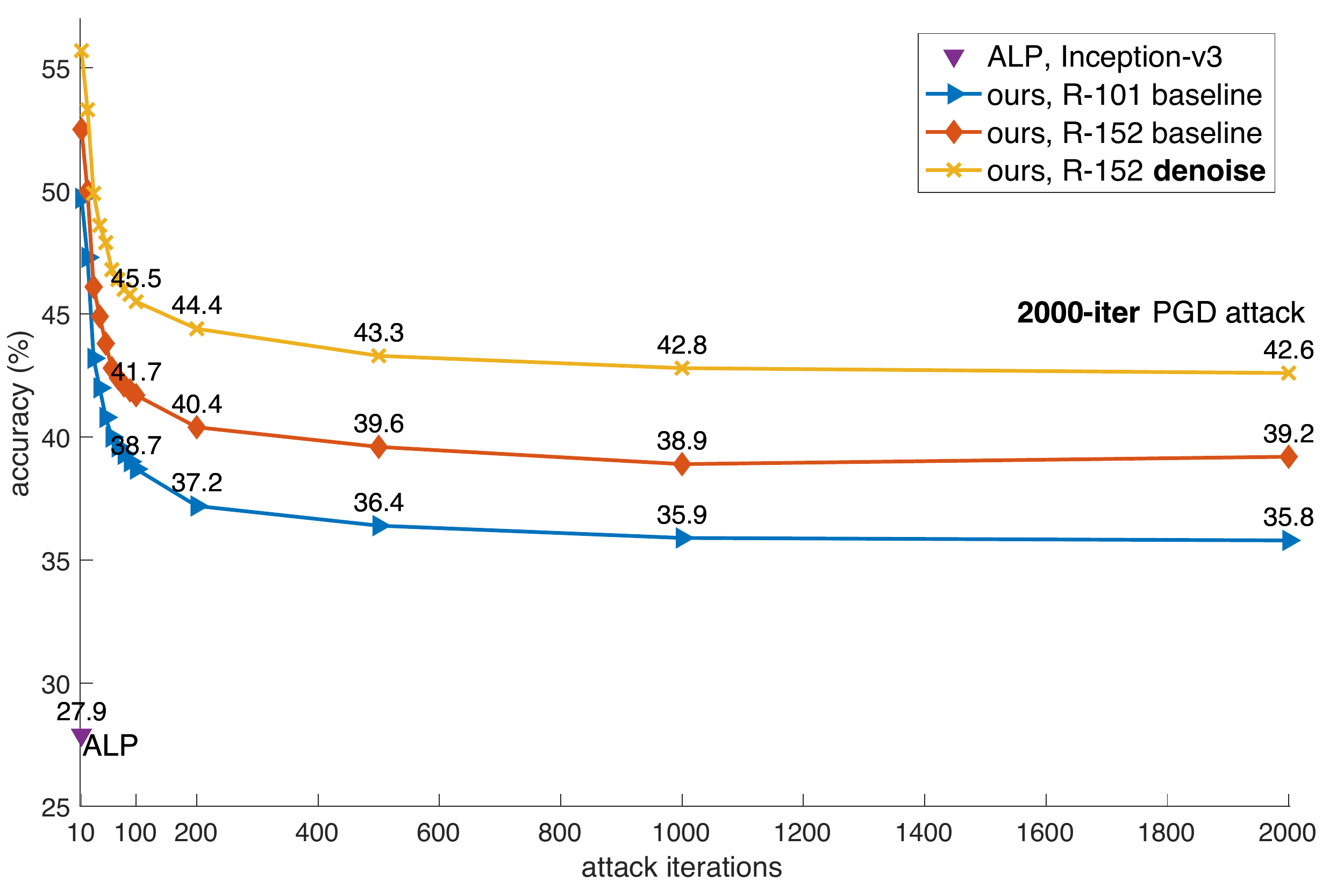}
\includegraphics[height=6.1cm]{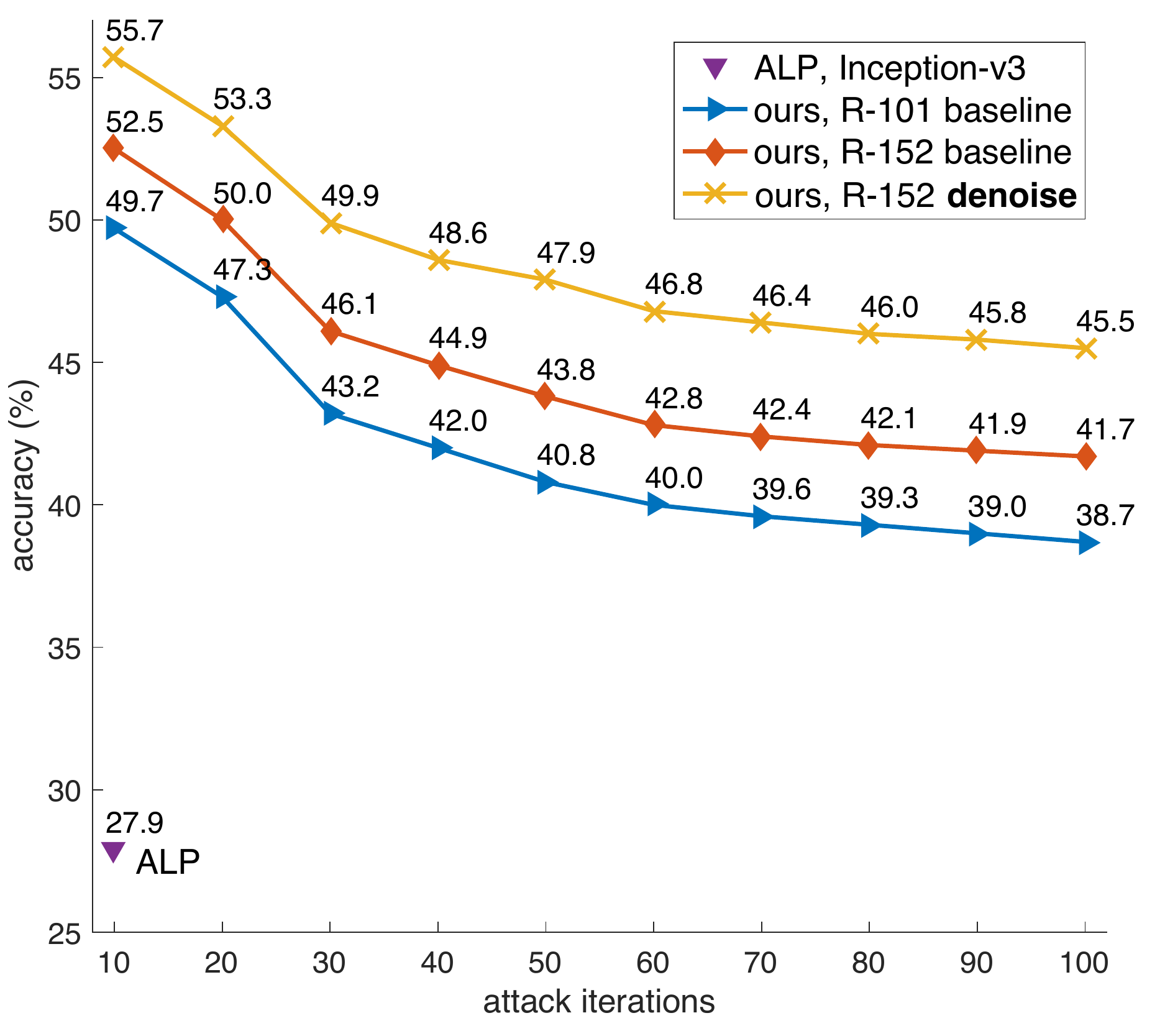}
}
\vspace{-1.5em}
\caption{\textbf{Defense against white-box attacks on ImageNet}. The left plot shows results against a white-box PGD attacker with 10 to \textbf{2000} attack iterations. The right plot zooms in on the results with 10 to 100 attack iterations.
The maximum perturbation is $\epsilon$ $=$ 16.
}
\label{fig:accuracy_baseline} 
\vspace{-.5em}
\end{figure*}

\paragraph{Main results.} Figure~\ref{fig:accuracy_baseline} shows the main results.
We first compare with ALP \cite{Kannan2018}, the previous state-of-the-art. ALP was evaluated under 10-iteration PGD attack in \cite{Kannan2018}, on Inception-v3 \cite{Szegedy2016a}. It achieves 27.9\% accuracy on ImageNet validation images (Figure~\ref{fig:accuracy_baseline}, purple triangle).

ResNet-101 and ResNet-152 in Figure~\ref{fig:accuracy_baseline} are our baseline models (\emph{without} any denoising blocks) trained using our adversarial training implementation. Even with the lower-capacity model of R-101, our baseline is very strong --- it has 49.7\% accuracy under 10-iteration PGD attacks, considerably better than the ALP result. This shows that our adversarial training system is solid; we note that the comparison with ALP is on the system-level as they differ in other aspects (backbone networks, implementations, \etc). 

``R-152, denoise'' in Figure~\ref{fig:accuracy_baseline} is our model of ResNet-152 with four denoising blocks added. Here we show the best-performing version (non-local with Gaussian), which we ablate next.  There is a consistent performance improvement introduced by the denoising blocks. Under the 10-iteration PGD attack, it improves the accuracy of ResNet-152 baseline by \textbf{3.2\%} from 52.5\% to 55.7\% (Figure~\ref{fig:accuracy_baseline}, right).

Our results are robust even under \textbf{2000-iteration} PGD attacks.
To our knowledge, such a strong attack has \emph{not} been previously explored on ImageNet.
ALP \cite{Kannan2018} was only evaluated against 10-iteration PGD attacks (Figure~\ref{fig:accuracy_baseline}), and its claimed robustness is subject to controversy \cite{Engstrom2018}.
Against 2000-iteration PGD attacks, our ResNet-152 baseline has 39.2\% accuracy, and its denoising counterpart is \textbf{3.4\%} better, achieving 42.6\%. We also observe that the attacker performance diminishes with 1000$\app$2000 attack iterations.

We note that in this \emph{white-box} setting, the attackers can iteratively back-propagate \emph{through} the denoising blocks and create adversarial perturbations that are tailored to the denoisers. 
Recent work \cite{Athalye2018} reports that \emph{pixel} denoising methods can be circumvented by attackers in the white-box settings. 
By contrast, feature denoising leads to consistent improvements in white-box settings, suggesting that \emph{feature} denoising blocks make it more difficult to fool networks.

\begin{figure}[tp!]
\centering
\vspace{-.5em}
\includegraphics[width=1.0\linewidth]{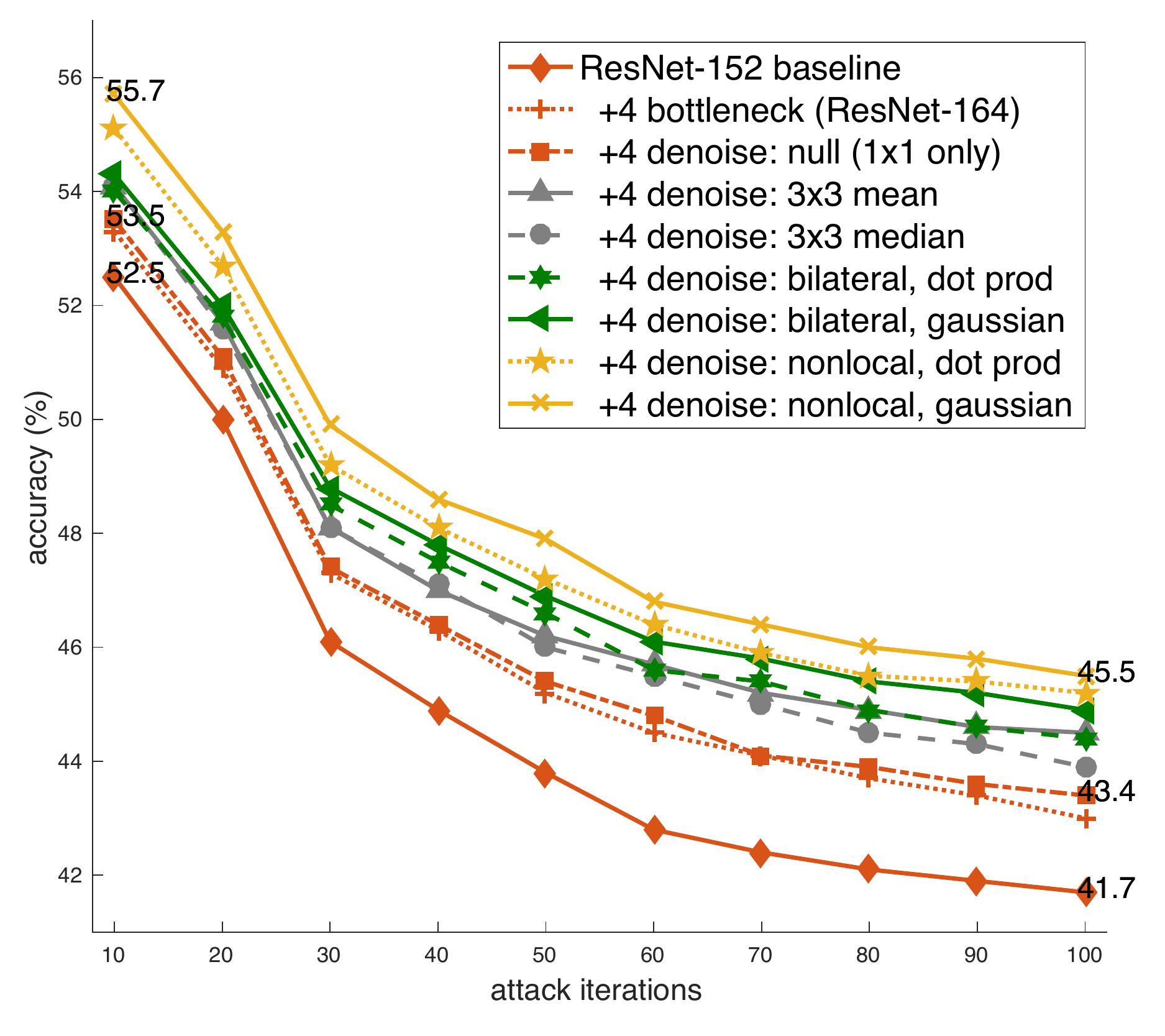}
\vspace{-1.8em}
\caption{\textbf{Ablation: denoising variants} for defending against \emph{white-box} attacks on ImageNet.
On the ResNet-152 baseline, all other models add \emph{4 blocks} to it.
The attacker is PGD under different attack iterations, with $\epsilon$ $=$ 16.
All denoising models are better than the R-152 baseline and the ``null" version.
}
\label{fig:accuracy_ablation}
\vspace{-1.5em}
\end{figure}

\paragraph{Variants of denoising operations.} Next, we evaluate variants of denoising operations in Sec.~\ref{sec:denoise}. In these ablations, we add blocks of different kinds to baseline ResNet-152.

We consider the following denoising operations: 3$\times$3 \emph{mean filtering}, 3$\times$3 \emph{median filtering}, 3$\times$3 \emph{bilateral filtering} (Eqn.~(\ref{eq:local})), and \emph{non-local filtering}.
In our ablation study, we further consider a ``\emph{null}'' version of the denoising block: the block in Figure~\ref{fig:block} becomes trivially a residual block with a single 1$\times$1 convolution.
Further, we also compare with adding 4 standard \emph{bottleneck} \cite{He2016} blocks --- essentially, ResNet-164.
All models are trained by adversarial training.
Figure~\ref{fig:accuracy_ablation} shows the white-box attacks results; for simplicity, we show PGD attacker with up to 100 attack iterations in this ablation.

\emph{All} of these denoising operations have better accuracy than: (i) ResNet-152 baseline, (ii) adding 4 standard bottleneck blocks, and (iii) adding 4 ``null'' denoising blocks. It is worth noticing that the 1$\times$1 null version has \emph{the exact same number of extra parameters} as the mean filtering, median filtering, and bilateral/non-local filtering's dot product versions (which have no embedding).
The null version is worse than all of them (Figure~\ref{fig:accuracy_ablation}).
Also, while adding standard bottleneck blocks is helpful, adding denoising blocks of \emph{any} version is more accurate.
These results suggest that the extra parameters are \emph{not} the main reason for our accuracy improvements; feature denoising appears to be a general approach particularly useful for adversarial robustness.

Our best-performing model is given by the non-local (Gaussian) version, which we use by default in other parts of the paper unless noted. Interestingly, this Gaussian version is just marginally better than the dot product version.

\renewcommand\arraystretch{1.02}
\setlength{\tabcolsep}{12pt}
\begin{table}[t]
\centering
\resizebox{.75\linewidth}{!}{
\small
\begin{tabular}{l|cc}
attack iterations               & 10 & 100 \\
\shline
non-local, Gaussian  & 55.7 & 45.5 \\
\hline
removing 1$\times$1  & 52.1 & 36.8 \\
removing residual  & NaN & NaN \\
\end{tabular}
}
\vspace{.5em}
\caption{\textbf{Ablation: denoising block design} for defending against \emph{white-box} attacks on ImageNet. Our networks have four (Gaussian) non-local means denoising blocks. We indicate the performance of models we were unable to train by ``NaN''. 
}
\label{tab:design}
\vspace{-.5em}
\end{table}

\paragraph{Design decisions of the denoising block.} The denoising block in Figure~\ref{fig:block} has a 1$\times$1 layer and a residual connection. Although both components do not perform denoising, they are important for the denoising blocks to work well. Next, we ablate the behavior of the 1$\times$1 and residual connection.

This ablation is in Table~\ref{tab:design}.
We investigate ResNet-152 with four non-local, Gaussian denoising blocks.
All models are all trained by adversarial training.
When removing the 1$\times$1 convolution in the denoising block, the accuracy drops considerably --- \eg, decreasing from 45.5\% to 36.8\% under 100-iteration PGD attacks.
On the other hand, removing the residual connection makes training unstable, and its loss does not decrease in our adversarial training.

These results suggest that denoising features \emph{in itself} is not sufficient. As suppressing noise may also remove useful signals, it appears essential to properly combine the denoised features with the input features in denoising blocks.

\subsection{Against Black-Box Attacks} 

Next, we evaluate defending against \emph{black-box} attacks.
To have an unbiased yet challenging set of attackers, we study the \emph{5 best attackers} of the NIPS 2017 CAAD competition \cite{Kurakin2018}, for which code is publicly available.
We use the latest CAAD 2018 evaluation criterion, which we call ``all-or-nothing'': \emph{an image is considered correctly classified \underline{only if} the model correctly classifies \underline{all} adversarial versions of this image created by all attackers}. This is a challenging evaluation scenario for the defender. Following the CAAD black-box setting, the maximum perturbation for each pixel is $\epsilon=$~\textbf{32}, which also makes defense more difficult. 
Note that our models are trained with $\epsilon=16$.

Table~\ref{tab:blackbox} shows the results of defending against black-box attacks on ImageNet validation images. 
To highlight the difficulty of the new ``all-or-nothing" criterion, we find that the CAAD 2017 winner \cite{Liao2018} has only 0.04\% accuracy under this criterion. We find that it is mainly vulnerable to two of the five attackers\footnote{\fontsize{6pt}{1em}\selectfont \url{https://github.com/pfnet-research/nips17-adversarial-attack}}$^{,}$\footnote{\fontsize{6pt}{1em}\url{https://github.com/toshi-k/kaggle-nips-2017-adversarial-attack}}. If we remove these two attackers, \cite{Liao2018} has 13.4\% accuracy in the ``all-or-nothing" setting.

With the ``all-or-nothing" criterion, our ResNet-152 baseline has 43.1\% accuracy against all five attackers. This number suggests that a successful implementation of adversarial training is critical for adversarial robustness. 

\renewcommand\arraystretch{1.1}
\setlength{\tabcolsep}{3pt}
\begin{table}[t]
\centering
\resizebox{.85\linewidth}{!}{
\small
\begin{tabular}{l|c}
\multicolumn{1}{c|}{model}               & accuracy (\%)\\
\shline
CAAD 2017 winner  &  0.04 \\
\demph{CAAD 2017 winner, under 3 attackers}  &  \demph{13.4} \\
\hline
ours, R-152 baseline   &      43.1        \\
\quad +4 denoise: null (1$\times$1 only)   &         44.1               \\
\quad +4 denoise:  non-local, dot product   &        46.2                  \\
\quad +4 denoise:  non-local, Gaussian   &           \textbf{46.4}                 \\
\hline
\quad +all denoise: non-local, Gaussian  & \textbf{49.5} \\
\end{tabular}
}
\vspace{.7em}
\caption{\textbf{Defense against black-box attacks on ImageNet}. We show top-1 classification accuracy on the ImageNet validation set.
The attackers are the 5 best attackers in CAAD 2017. We adopt the CAAD 2018 ``\emph{all-or-nothing}'' criterion for defenders. 
The 2017 winner has 0.04\% accuracy under this strict criterion, and if we remove the 2 attackers that it is most vulnerable  to, it has 13.4\% accuracy under the 3 remaining attackers.
}
\label{tab:blackbox}
\end{table}

On top of our strong ResNet-152 baseline, adding four non-local denoising blocks improves the accuracy to 46.4\% (Table~\ref{tab:blackbox}). Interestingly, both the Gaussian and dot product versions perform similarly (46.4\% \vs 46.2\%), although the Gaussian version has more parameters due to its embedding. Furthermore, the null version has 44.1\% accuracy --- this is worse than the non-local, dot product version, even though they have the same number of parameters; this null version of 1$\times$1 is 1.0\% better than the ResNet-152 baseline.

We have also studied the local variants of denoising blocks, including mean, median, and bilateral filters. They have 43.6\% $\app$ 44.4\% accuracy in this black-box setting. Their results are not convincingly better than the null version's results. This suggests that non-local denoising is more important than local denoising for robustness against these black-box attackers.

\paragraph{Pushing the envelope.} To examine the potential of our model, we add denoising blocks to \emph{all} residual blocks (one denoising block after each residual block) in ResNet-152.
We only study the non-local Gaussian version here. 
To make training feasible, we use the sub-sampling trick in \cite{Wang2018}: the feature map of $x_j$ in Eqn. (\ref{eq:nonlocal}) is subsampled (by a 2$\times$2 max pooling) when performing the non-local means, noting that the feature map of $x_i$ is still full-sized. We only use sub-sampling in this case. It achieves a number of 49.5\%. This is \textbf{6.4\%} better than the ResNet-152 baseline's 43.1\%, under the black-box setting (Table~\ref{tab:blackbox}).

\paragraph{CAAD 2018 challenge results.} Finally, we report the results from the latest CAAD 2018 competition. The 2018 defense track adopts the ``all-or-nothing'' criterion mentioned above --- in this case, every defense entry needs to defend against 48 \emph{unknown} attackers submitted to the same challenge (in contrast to 5 attackers in our above black-box ablation). 
The test data is a secret, ImageNet-like dataset.
The maximum perturbation for each pixel is $\epsilon=32$.

\begin{figure}[t]
\centering
\includegraphics[width=.83\linewidth]{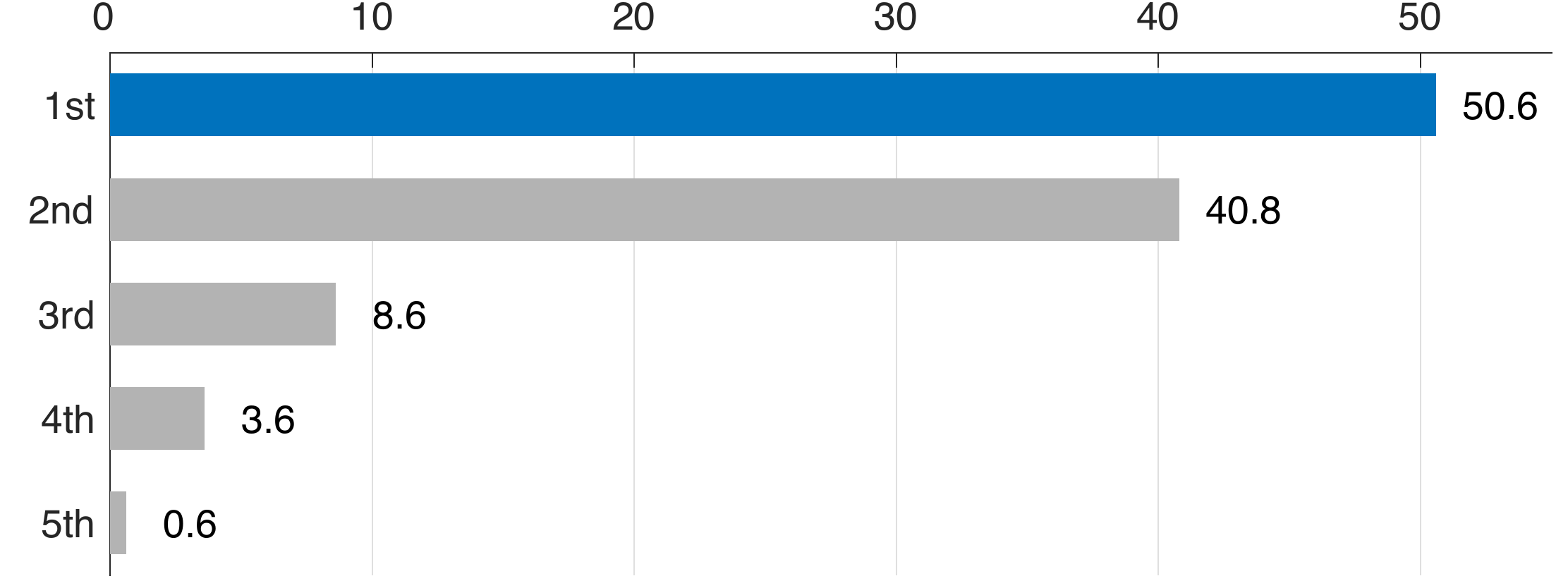}
\caption{\textbf{CAAD 2018 results of the adversarial defense track}.
The first-place entry is based on our method. We only show the 5 winning submissions here, out of more than 20 submissions.}
\label{fig:caad2018}
\vspace{-1.5em}
\end{figure}

Figure~\ref{fig:caad2018} shows the accuracy of the 5 best entries in the CAAD 2018 defense track. The winning entry, shown in the blue bar, was based on our method by using a ResNeXt-101-32$\times$8 backbone \cite{Xie2017} with non-local denoising blocks added to all residual blocks. This entry only uses \emph{single-crop, single-model} testing. It achieves \textbf{50.6\%} accuracy against 48 unknown attackers. This is $\app$\textbf{10\%} absolute (20\% relative) better than the second place's 40.8\% accuracy. 

We also reported the white-box performance of this winning entry on ImageNet. Under 10-iteration PGD attacks and 100-iteration PGD attacks, it achieves 56.0\% accuracy and 40.4\% accuracy, respectively. These results are slightly worse than the robustness of ResNet-152 based models reported in Section \ref{subsec; white-box exp}. We note that this white-box robustness comparison is on the system-level as the winning entry was trained with a slightly different parameter setting. 

We emphasize that the CAAD 2018 defense task is very challenging because of the ``all-or-nothing'' criterion and many unknown (potentially new state-of-the-art) attackers. Actually, except for the two leading teams, all others have $<$10\% accuracy and many of them have $<$1\% accuracy. This highlights the significance of our 50.6\% accuracy.

\subsection{Denoising Blocks in Non-Adversarial Settings}

Thus far we have been focusing on denoising blocks for improving adversarial defense. Because our denoising blocks are components of the convolutional networks, these networks can also be trained without adversarial training for the classification of ``clean'' images (\ie, the original ImageNet dataset task). We believe that studying the non-adversarial setting can help us better understand the behavior of denoising blocks.

Table~\ref{tab:clean} presents the clean image performance of models that were not adversarially trained. We compare the baseline R-152, adding standard bottleneck blocks, adding ``null" (1$\times$1) denoising blocks, and adding denoising blocks of various types. In the clean setting, these denoising blocks have \emph{no obvious advantage} over the baseline R-152, adding standard bottleneck blocks, or adding ``null" denoising blocks. Actually, all results are in the range of about $\pm$0.2\% of the baseline R-152's result --- which have \emph{no significant difference} if we also consider the natural variance between separate training runs of the same model (see baseline R-152 in Table~\ref{tab:clean}).

We also find that adding non-local denoising blocks to the shallower ResNet-50 can moderately improve accuracy by 0.7\% in the non-adversarial setting, but doing so on ResNet-152 has diminishing gain. This, however, is not the case for the adversarial images.

These results suggest that the denoising blocks could have special advantages in settings that require adversarial robustness. This observation matches our intuition that denoising blocks are designed to reduce feature noise, which only appears when classifying adversarial images.

Finally, we report that our ResNet-152 baseline with \emph{adversarial} training has 62.32\% accuracy when tested on \emph{clean} images, whereas its counterpart with ``clean'' training obtains 78.91\%. 
For the denoising version (non-local, Gaussian), the accuracy of an adversarially trained network is 65.30\% on clean images, whereas its cleanly trained counterpart obtains 79.08\%. This tradeoff between adversarial and clean training has been observed before (\eg, in \cite{Tsipras2018}); we expect this tradeoff to be the subject of future research.

\renewcommand\arraystretch{1.05}
\setlength{\tabcolsep}{5pt}
\begin{table}[t]
\centering
\resizebox{.75\linewidth}{!}{
\small
\begin{tabular}{l|c}
model & \multicolumn{1}{c}{accuracy (\%)} \\
\shline
R-152 baseline  &                                  78.91  \\ 
R-152 baseline, run 2	&                        +0.05 \\  
R-152 baseline, run 3 	&                         -0.04 \\  
\hline
+4 bottleneck (R-164) 	&                         +0.13 \\  
+4 denoise: null (1$\times$1 only)  &     +0.15 \\ 
+4 denoise: 3$\times$3 mean filter &     +0.01 \\ 
+4 denoise: 3$\times$3 median filter &   -0.12 \\ 
+4 denoise: bilateral, Gaussian   &         +0.15 \\ 
+4 denoise: non-local, Gaussian  &        +0.17 \\ 
\end{tabular}
}
\vspace{.5em}
\caption{\textbf{Accuracy on clean images} in the ImageNet validation set when trained on clean images. 
All numbers except the first row are reported as the accuracy difference comparing with the first R-152 baseline result.
For R-152, we run training 3 times independently, to show the natural random variation of the same architecture.
All denoising models show \emph{no significant difference}, and are within $\pm$0.2\% of the baseline R-152's result.
}
\label{tab:clean}
\vspace{-1em}
\end{table}

\section{Conclusion}

Motivated by the noisy appearance of feature maps from adversarial images, we have demonstrated the potential of feature denoising for improving the adversarial robustness of convolutional networks. Interestingly, our study suggests that there are certain \emph{architecture designs} (\emph{viz.}, denoising blocks) that are particularly good for adversarial robustness, even though they do not lead to accuracy improvements compared to baseline models in ``clean'' training and testing scenarios. When combined with adversarial training, these particular architecture designs may be more appropriate for modeling the underlying distribution of adversarial images. We hope our work will encourage researchers to start designing convolutional network architectures that have ``innate'' adversarial robustness.

{\small
\bibliographystyle{ieee}
\bibliography{defeat}
}

\end{document}